%% file: paper.tex
\documentclass[times,twocolumn,final]{elsarticle}

\usepackage{medima}
\usepackage{framed,multirow}
\usepackage{multicol} 

\usepackage{amssymb}
\usepackage{latexsym}

\usepackage{url}
\usepackage{xcolor}
\usepackage{hyperref}

\usepackage{graphicx}
\usepackage{enumitem}
\usepackage{amsmath}
\usepackage{makecell} 

\input{definitions}
\def\iqs{iQ-solutions\texttrademark\@\xspace}
\def\DiceThresh{Minor Boundary Adjustment for Threshold\@\xspace}
\def\DT{MinBAT\@\xspace}
\def\PatchSelect{ROI-based Expanded Patch Selection\@\xspace}
\def\PS{REPS\@\xspace}
\def\MSData{SNAC-MS\@\xspace} 
\def\BrData{SNAC-Brain\@\xspace}

\def\thickline{\Xhline{3\arrayrulewidth}}

\journal{Medical Image Analysis (Under Review)}

\begin{document}

\verso{D. Wang, P, Liu, H. Wang~\etal}

\begin{frontmatter}

\title{How Much Data are Enough? Investigating Dataset Requirements for Patch-Based Brain MRI Segmentation Tasks}
%

\author[1,2]{Dongang \snm{Wang}}
\ead{dongang.wang@sydney.edu.au}
\author[3]{Peilin \snm{Liu}}
\author[2]{Hengrui \snm{Wang}}
\author[1,4]{Heidi \snm{Beadnall}}
\author[1,2]{Kain \snm{Kyle}}
\author[2]{Linda \snm{Ly}}
\author[1]{Mariano \snm{Cabezas}}
\author[1,2]{Geng \snm{Zhan}}
\author[5,6]{Ryan \snm{Sullivan}}
\author[1,4]{Weidong \snm{Cai}}
\author[7]{Wanli \snm{Ouyang}}
\author[1,8,9]{Fernando \snm{Calamante}}
\author[1,2,4]{Michael \snm{Barnett}\fnref{fn1}}
\author[1,2]{Chenyu \snm{Wang}\corref{cor1}\fnref{fn1}}
\ead{chenyu.wang@sydney.edu.au}

\address[1]{Brain and Mind Centre, The University of Sydney, Camperdown, NSW 2050, Australia}
\address[2]{Sydney Neuroimaging Analysis Centre, 94 Mallett Street, Camperdown, NSW 2050, Australia}
\address[3]{School of Mathematics and Statistics, The University of Sydney, Camperdown, NSW 2006, Australia}
\address[4]{Royal Prince Alfred Hospital, Camperdown, NSW 2050, Australia}
\address[5]{School of Computer Science, The University of Sydney, Camperdown, NSW 2006, Australia}
\address[6]{Australian Imaging Service, Camperdown, NSW 2006, Australia}
\address[7]{Shanghai Artificial Intelligence Laboratory, China}
\address[8]{School of Biomedical Engineering, The University of Sydney, Camperdown, NSW 2006, Australia}
\address[9]{Sydney Imaging, The University of Sydney, Camperdown, NSW 2006, Australia}
            
\cortext[cor1]{Corresponding author}

\fntext[fn1]{Equal contributed senior authors}


\begin{abstract}
Training deep neural networks reliably requires access to large-scale datasets. However, obtaining such datasets can be challenging, especially in the context of neuroimaging analysis tasks, where the cost associated with image acquisition and annotation can be prohibitive. To mitigate both the time and financial costs associated with model development, a clear understanding of the amount of data required to train a satisfactory model is crucial.

This paper focuses on an early stage phase of deep learning research, prior to model development, and proposes a strategic framework for estimating the amount of annotated data required to train patch-based segmentation networks. This framework includes the establishment of performance expectations using a novel \DiceThresh (\DT) method, and standardizing patch selection through the \PatchSelect (\PS) method.

Our experiments demonstrate that tasks involving regions of interest (ROIs) with different sizes or shapes may yield variably acceptable Dice Similarity Coefficient (DSC) scores. By setting an acceptable DSC as the target, the required amount of training data can be estimated and even predicted as data accumulates.
This approach could assist researchers and engineers in estimating the cost associated with data collection and annotation when defining a new segmentation task based on deep neural networks, ultimately contributing to their efficient translation to real-world applications.

\end{abstract}

\begin{keyword}
\KWD \\
Data requirements \sep \\
Patch-based segmentation \sep \\
Dice similarity coefficient
\end{keyword}

\end{frontmatter}
%
%
\section{Introduction}
Numerous deep learning algorithms have recently been proposed to analyze patterns in various medical imaging modalities by solving challenges in classification, object detection, and segmentation tasks~\citep{litjens2017survey}. 
The proliferation of deep neural network models is closely related to the increasing availability of annotated data. 
A long-held consensus is that larger training datasets are crucial for training deeper network models~\citep{sun2017revisiting}; and that the performance and robustness of these algorithms can be boosted with more data to mitigate overfitting~\citep{zhang2021understanding}.
For example, the widely used ImageNet dataset contains more than 14 million mid-resolution images~\citep{russakovsky2015imagenet} and covers 1000 object categories, serving as a foundational resource to provide pre-trained models for downstream tasks. Recent breakthroughs, exemplified by models like GPT-3~\citep{brown2020language} and Segment Anything Model~\citep{Kirillov_2023_ICCV}, have been pre-trained on even larger datasets, with 570GB of sentences, and more than 11 million high-resolution images labeled with 1 billion masks, respectively. 

However, the process of collecting and annotating large-scale datasets, particularly in the realm of the neuroimaging domain, requires specialty expertise and is both time-consuming and financially demanding.

As an example, academic multiple sclerosis (MS) clinical centers in Australia refer approximately 100-150 patients per month for MRI scans. However, only a fraction of these scans can be included in the development of AI models, considering patient consent, image quality, duplicate (\ie, non-unique) scans, and acquisition protocol requirements.
Semi-automated lesion annotation of these scans using traditional neuroimaging tools requires between 10-40 minutes of work by a trained neuroimaging analyst, depending on the lesion burden.
To meet the rigorous standard imposed by clinical trials and ensure the reliability and quality of the annotated lesion masks, collaborative efforts from multiple junior and senior neuroimaging analysts are required for quality checks and annotation calibration.

While MS is a relatively common disorder, it may prove even more difficult to collect data for rare brain diseases or conditions, such as Amyloid Related Imaging Abnormalities (ARIA)~\citep{barkhof2013mri}, that occur only in less than 8\% of cases of Alzheimer's disease analyzed by~\citep{jeong2022incidence}.

Consequently, the expectation that large-scale annotated datasets will be readily available to train robust deep learning models for neuroimaging applications is often unrealistic. 
A clear understanding of the data size required to achieve acceptable, rather than perfect, model performance, is a balanced and in most cases advantageous approach for both academic and industry pursuits.

Exploration of the dataset requirements for deep learning has to date primarily focused on providing empirical analysis. In~\citep{hestness2017deep}, extensive experiments were conducted across various tasks, including language translation, image classification, and speech recognition. The authors proposed that model performance may follow a power-law function as the volume of training data increases, ultimately reaching an upper bound determined by inherent errors in the training data and annotations.
In~\citep{mahmood2022much}, the authors conducted experiments to predict data requirements based on several monotonically increasing regression functions, emphasizing that with an accumulation of data and experiments, the margin of error in predicting the model performance with higher data amount can be reduced. 
In~\citep{tejero2023full}, an approach based on Gaussian process was designed by considering both model performance and financial budget to determine the optimal number of cases for both classification and segmentation tasks.



Translation of this research to medical images is hampered by the unique challenges posed by 3D segmentation tasks, which are increasingly important for neuroimaging analysis, such as monitoring lesion activity (namely, the development of new or enlarging lesions) in patients with MS~\citep{ma2022multiple}, or measuring brain structural change in neurodegenerative disease~\citep{zhan17learning}. 
Two specific questions, therefore, must be answered to overcome these roadblocks: first, `\textit{how should the expected performance for different tasks be determined?}' and, second, `\textit{how should existing prediction methods be transformed for application to segmentation tasks in 3D images?}'

In contrast to research endeavors that target perfect performance with given data, the real-world application of algorithms may not necessarily demand exceedingly high performance, especially when achieving such ``extra'' capability may require substantively greater resources, which compromise its practical feasibility. Therefore, it becomes important to define a performance target to determine whether the available data are sufficient to train a robust model in real practice. 
However, the target may not be identical for different segmentation tasks.
For example, in the case of MS lesion segmentation, it is exceptionally challenging to attain a Dice Similarity Coefficient (DSC) exceeding 0.70 by present methods~\citep{ma2022multiple,kamraoui2022deeplesionbrain}. 
Similarly, in~\citep{carass2020evaluating}, DSC scores were computed for MS lesion segmentation under both inter-rater and inter-algorithm scenarios. Notably, it was found that different raters had diverging opinions on the same group of cases, resulting in DSC values ranging from 0.65 to 0.67 when compared to the consensus. As discovered by~\citep{kamraoui2022deeplesionbrain}, the same method trained on different datasets could hardly have a consistency DSC higher than 0.7. 
These values, while slightly higher than those achieved by analyzed algorithms in~\citep{ma2022multiple}, are far from the empirically ideal DSC score (\eg, $>$0.9) that can be observed in other segmentation tasks, such as brain extraction.
For example, brain extraction algorithms on MRI could attain a DSC of 0.99~\citep{teng2023automated}, and even in fetal neurosonography, in which images are considerably more ``blurred'' than MRI, a DSC of 0.94 is achievable~\citep{moser2022bean}. 

The expectations regarding DSC performance can therefore diverge significantly in different segmentation tasks with different sizes of regions of interest (ROIs), and establishing a universal DSC performance target to guide data and annotation preparation for all segmentation tasks is difficult and potentially misleading.

Furthermore, when training and evaluating 3D segmentation models for medical imaging, it is common to utilize 3D patches due to resource constraints~\citep{isensee2018nnu}. 
Given the diversity of ROIs in neuroimaging segmentation tasks with respect to sizes, resolution, and distribution, various patch selection methods can be applied~\citep{tang2021high} to generate patches for training, and random patches are widely used as a key data augmentation technique in practice~\citep{isensee2018nnu}. However, in applications where random numbers of patches are selected from each training case, it is challenging to determine the contribution of each individual case.
This difference hinders the application of previous data requirement estimation methods to patch-based medical image segmentation tasks, as the required number of patches predicted by these methods is not directly indicative of the required number of cases.

The research efforts described in this paper focus on the development of a strategic framework for estimating the amount of annotated data required to train patch-based segmentation networks.
Specifically, to establish the minimum target of the evaluation metric for a specific model, we introduce a \DiceThresh (\DT) method, a dynamic approach to set specific acceptable values of DSC according to the ROI of each task. 
To standardize the contribution of cases to the training process, we additionally present a \PatchSelect (\PS) strategy to maintain model performance, while equalizing the impact of each newly included case. 
Through learning processes that utilize \DT and \PS as data accumulate, we successfully demonstrate clear learning curves that illustrate the relationship between model performance and the number of training cases.
This experimental work supports efforts to assess task-specific data requirements and ultimately informs the quantity of data needed to sufficiently, albeit imperfectly, train a segmentation model.

In summary, the contributions described in this paper are three-fold:
\begin{enumerate}
    \item By revisiting the definition of DSC and aligning this with real-world clinical requirements, we identify the key factor to influence DSC scores and propose a \DiceThresh (\DT) strategy based on the Markov process to determine expected levels of DSC for each segmentation task.
    \item We introduce a \PatchSelect (\PS) strategy to prepare patches from cases that are able to simultaneously incorporate data augmentation techniques to maintain model performance and standardize the contribution of each training case.
    \item Experiments were conducted on three widely studied brain-related segmentation tasks, including brain extraction, tumor segmentation, and MS lesion segmentation, with different ROI sizes and shapes. The effectiveness of the proposed pipeline has been verified by visualizing learning curves and the number of required cases can be estimated and predicted based on a limited number of collected cases.
    \item Relationships between the number of cases, the number of ROIs, and the achievable performance have been discovered through both theoretical derivatives and experiments. The findings potentially assist the design of experiments and evaluation of algorithms.
\end{enumerate}

\section{Methods and Materials}
\subsection{\DiceThresh (\DT) by Markov Process}\label{sec4:markov}
In clinical applications of segmentation tasks, achieving a 100\% accurate DSC score is exceptionally challenging. This difficulty is intricately related to the fundamental definition of the DSC score and its characteristics.

\begin{figure}[t]
    \centering
    \includegraphics[width=0.5\textwidth]{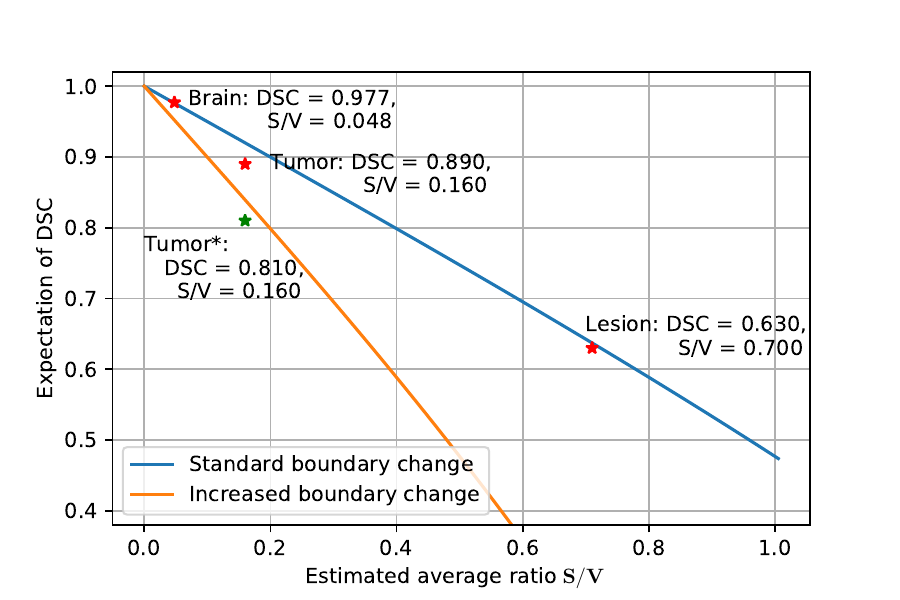}
    \caption{Visualization of expectation of DSC with respect to ratio $\S/\V$. The proposed DSC thresholds of tasks are displayed as stars. The blue curve represents Equation~\ref{eq4:integral} under the assumption of minimal mask variation near the boundary, and red stars represent the estimation results with these standard boundary changes. The orange line represents a specific scenario with increased boundary changes, exemplified by the tumor (depicted by the green star). More details and discussions can be found in Section~\ref{sec4:exp-thresh} and Section~\ref{sec4:hyper}.}
    \label{fig4:integral}
\end{figure}

The DSC score is calculated by quantifying the volume overlap between the prediction and the target, divided by the average volume of the two. 
Specifically for one single image, the ground-truth mask is represented by the volume of $\V$ and the corresponding prediction mask as $\V'$. Compared with the ground truth $\V$, $\V'$ could contain $\Delta \V_1$ additional volumes and may also lack $\Delta \V_2$ volumes. The DSC score can be calculated as follows:

\begin{equation}
\begin{aligned}
    \text{DSC} &= \frac{2 \times (\V - \Delta \V_2)}{(\V + \V')} = \frac{2 \times (\V - \Delta \V_2)}{(2 \times \V + \Delta \V_1 - \Delta \V_2)}\\
    &= \frac{1- \Delta \V_2/\V}{1 - (\Delta \V_2 - \Delta \V_1)/2\V}. 
\end{aligned}\label{eq4:dice}
\end{equation}

When we consider a good prediction $\V'$, the volume changes $\Delta \V_1$ and $\Delta \V_2$ are primarily at the boundary of the masks and are associated with random variables $\mu_1$ and $\mu_2$ as the proportion of the volume changes along the mask border. 
In cases where the volume changes at the boundaries are relatively small, it can be reasonably assumed that $\Delta \V_1$ and $\Delta \V_2$ are proportional to the surface area $\S$ of the ROI. Therefore, Equation~\ref{eq4:dice} can be reformulated as follows:

\begin{equation}
    \text{DSC} = \frac{1- \mu_2 \cdot \S/\V}{1 - (\mu_2/2 - \mu_1/2) \cdot \S/\V}. 
\label{eq4:factor}
\end{equation}

From Equation~\ref{eq4:factor}, it is obvious that when any of the parameters, $\S/\V$ (the ratio of ROI surface area to ROI volume), $\mu_2$ (the probability of missed boundary voxels), and $\mu_1$ (the probability of additional boundary voxels), become higher, the achievable DSC score tends to be lower. As calculated in~\nameref{sec4:app}, DSC can be estimated as:

\begin{equation}
    \mathbb{E}(\text{DSC}) \approx -0.0279(\S/\V)^3 + 0.0063(\S/\V)^2 - 0.5016(\S/\V) + 1,
\label{eq4:integral}
\end{equation}
which shows that the DSC is approximately linear to $\S/\V$ of the task. The visualization of Equation~\ref{eq4:integral} is shown in Figure~\ref{fig4:integral}.

In the applicable scenario, it is possible that even with an acceptable model design and perfect data collection and annotation so that both $\mu_1$ and $\mu_2$ are minimized, the final DSC scores may not be high enough according to $\S/\V$. 
Empirically, for some tasks, DSC scores are hard to reach higher than a performance threshold (for example, 0.7 for MS lesion segmentation, and 0.9 for brain tumor core segmentation), and most of the errors are simply caused by small boundary-wise volume changes that may not hold clinical significance.

\begin{figure}[t]
    \centering
    \includegraphics[width=0.5\textwidth]{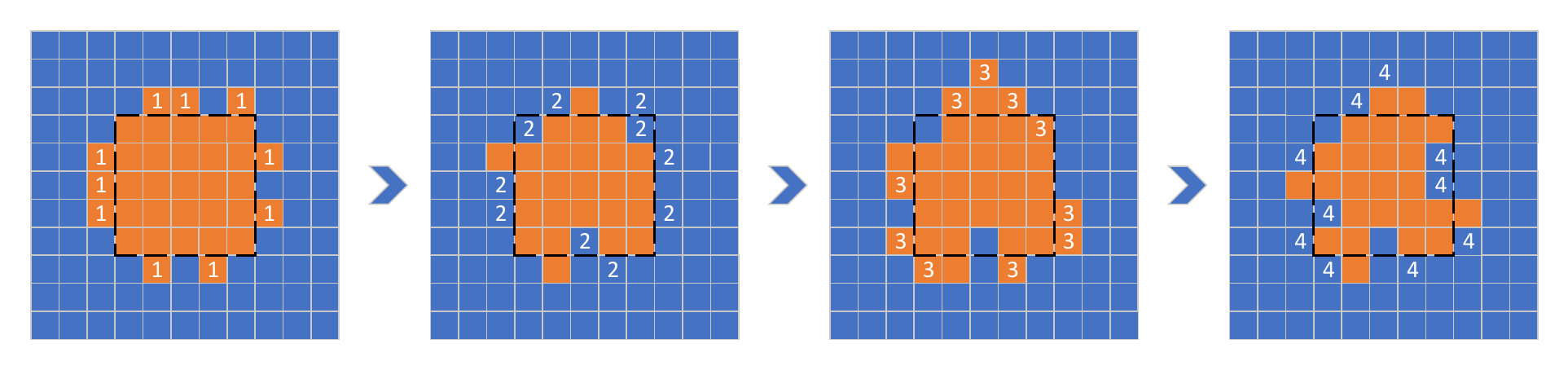}
    \caption{Example of the \DT process to generate random changes at the boundaries of ROI. Numbers 1 and 3 represent two steps of random dilation, and numbers 2 and 4 represent two steps of random erosion. The black dashed square marks the original mask. In our settings, the random probabilities for morphological changes were set to 0.5. In our experiments for real masks, such a process was conducted on 3D boundaries.}
    \label{fig4:markov}
\end{figure}

Given this insight, it becomes feasible to simulate the expected acceptable DSC for specific segmentation tasks based solely on annotated masks by estimating the average $\S/\V$ of each dataset. 
We followed the Markov process method proposed in~\citep{yao2023learning} (as shown in Figure~\ref{fig4:markov}) to calculate random masks of collected datasets by introducing small disturbances on the annotated ground truths controlled by random values $\mu_1$ and $\mu_2$ without changing $\S/\V$ substantially.
The modified masks should be used to calculate DSC scores by using the ground truth as the target. After multiple iterations of the Markov process on the original masks, a distribution of DSC scores of each case can be calculated, and the expected acceptable DSC score is derived as the expectation of this distribution. We defined this method as \DiceThresh (\DT), which is intended to find the target DSC score for a specific task, which can be further applied to determine the required amount of data.

\subsection{\PatchSelect (\PS)}\label{sec4:reps}
Another challenge of predicting the required data size for neuroimaging segmentation tasks arises from the inherent characteristics of 3D image volumes, which often consist of millions of effective voxels. To facilitate GPU calculation within the constraints of limited GPU memory, these images are usually downsampled~\citep{cciccek20163d} or divided into patches with smaller sizes~\citep{isensee2018nnu} before being fed into neural networks.
In previously published work~\citep{liu2023multiple,kao2020improving}, positive patches (\ie, patches that substantially contain voxels from ROIs) were randomly selected for training.

For particular tasks, such as MS lesion segmentation and tumor segmentation, the proportion of positive voxels in the entire image is generally low, and there is marked heterogeneity of ROI size across cases. As an example, the total MS lesions in each case could range from a few millimeters to more than one or two centimeters in diameter~\citep{filippi2019assessment}. 
Therefore, randomly selecting positive patches can bias patches toward certain ROIs and potentially overlook cases with smaller ROIs. 
While the method of random selection may yield good performance, random numbers and positions of patches from training cases make it challenging to accurately estimate the contribution of each case, ultimately complicating the estimation of the required number of cases. 

Moreover, previous work~\citep{hestness2017deep} has indicated that it is advisable to exclude data augmentation techniques when estimating data requirements, as they render the contributions of cases uncontrollable when patches are modified in each epoch. 
However, the exclusion of data augmentation is not suitable for patch-based neuroimaging segmentation tasks, primarily due to the adverse impact on model performance conferred by the limited number of positive cases or positive areas in training samples. 
Therefore, while uniform patch selection methods using sliding windows offer control over the number of patches, they should be improved to introduce patch modifications that enhance model performance.

\begin{figure}
    \centering
    \includegraphics[width=0.5\textwidth]{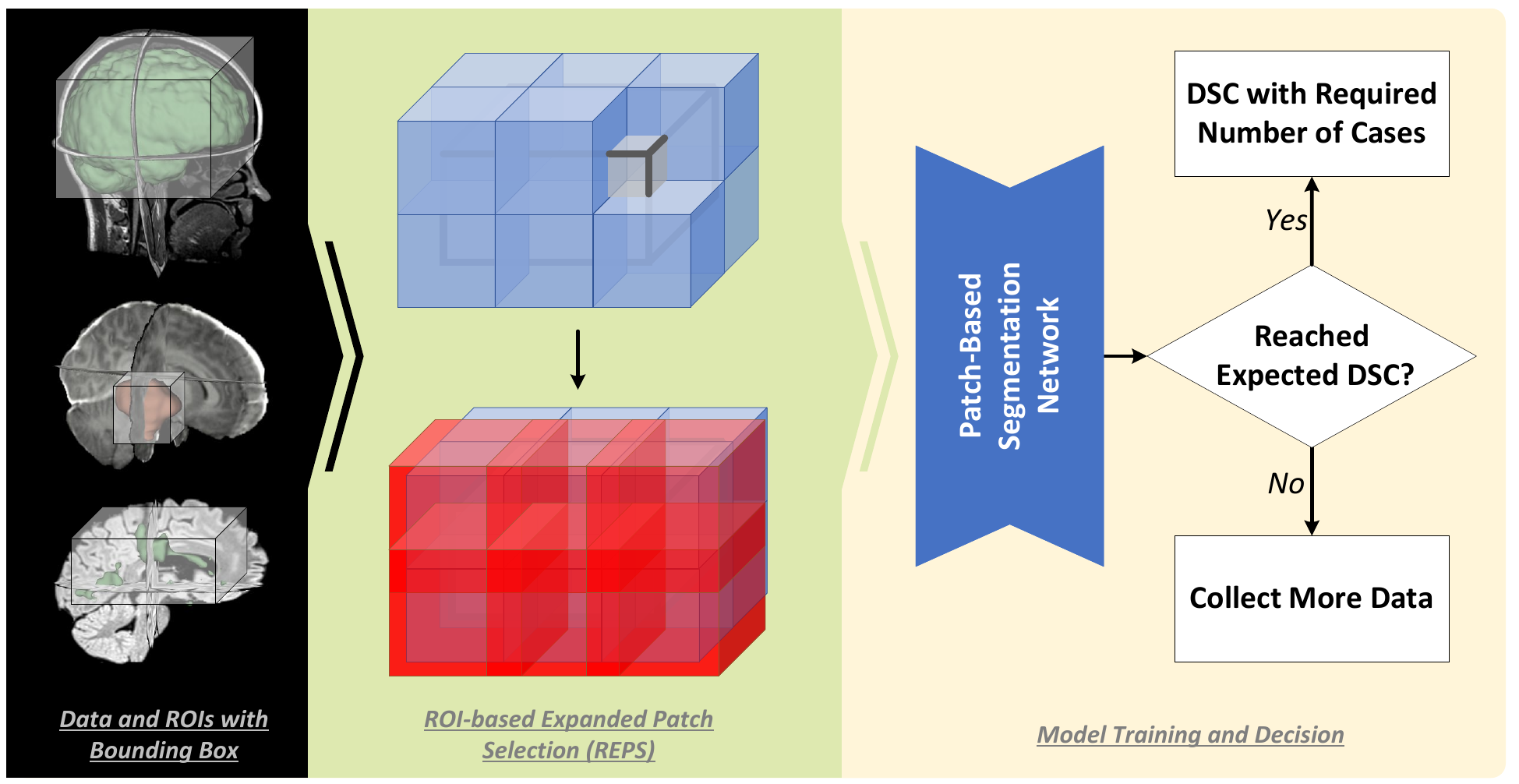}
    \caption{The pipeline of our \PatchSelect (\PS) and its application in selecting the expected number of cases. Three examples are shown to represent brain extraction (top), tumor extraction (middle), and lesion segmentation (bottom). Red patches are used in model training, where the patches with the same size as blue ones are selected randomly from each red patch in one epoch. }
    \label{fig4:pipeline}
\end{figure}

To overcome these challenges, we have developed the \PatchSelect (\PS) strategy for patch selection. This approach maintains randomness of data selection, preserving the model's training process and performance, while effectively controlling the number of patches. 
As illustrated in Figure~\ref{fig4:pipeline}, the first step involves selecting a bounding box that encompasses the entire ROI within the image (depicted as grey cubes in Figure~\ref{fig4:pipeline}). This bounding box encapsulates all positive annotations. 
Utilizing predefined patch size ($d_p$) and overlap size ($d_o$), we calculate the number of patches required to cover the entire valid bounding box (depicted as blue patches in Figure~\ref{fig4:pipeline}). 
Subsequently, these patches are further expanded using a predetermined boundary size ($d_b$), and the resulting expanded patches (depicted as red patches) are selected as input patches.

During the training process, patches of size $d_p^3$ are randomly selected from the input patches (with a size of $(d_p + 2d_b)^3$). 
By employing this \PS method, the number of patches is regulated consistently across all cases. While the number of patches may not be identical for each case, the differences are marginal, as the total patch count for each case is jointly determined by both the unique characteristics of the ROIs (which include the size, shape, and positions of ROIs, and may vary across cases) and the size of the input images (which is consistent across cases for the same training task). As a result, the randomness introduced by data augmentation is also maintained.

According to the pipeline shown in Figure~\ref{fig4:pipeline}, the methods \PS and \DT can be combined to evaluate whether a collected group of data is sufficient for neuroimaging segmentation tasks. 
In lieu of the conventional of ``\textit{collect as much data as possible}'', then ``\textit{train the model to maximal performance}'', which may be expensive and unnecessary for applications to meet usage requirements, we implemented a ``\textit{train as you go}'' strategy that trains models while collecting data, which has the capacity to infer data size requirements of the task by monitoring the delta in performance as data accumulate. If the target DSC calculated upon \DT is not reached, more data are needed. 
This approach represents an economic training framework, as the relationship between incremental data size and performance gains can be observed and communicated with project stakeholders for decision-making while providing a more accurate estimation of data size requirements as suggested by~\citep{mahmood2022much}.

\subsection{Experimental Datasets}
To fully illustrate the robustness of this framework, we conducted experiments to represent most segmentation challenges in neuroimaging, including brain extraction, tumor segmentation, and MS lesion segmentation, representing large, medium, and small ROIs. The experiments are based on separate datasets, including the public dataset BraTS~\citep{menze2014multimodal} and two internal datasets \BrData and \MSData.

Our internal dataset, \BrData, served as the basis for evaluating the brain extraction model. 
The original dataset comprised over 2000 brain MRI scans collected from a mix of healthy individuals and MS patients during the years 2012 to 2020 from three different scanners (a GE Discovery 3.0T, a Phillips Ingenia 3.0T, and a SIEMENS Skyra 3.0T). To ensure a balanced representation and to mitigate subject bias, we selected only one case per subject, with cases randomly chosen from all cases associated with a single subject's ID. This selection process yielded a final dataset of 749 subjects for training, with an additional 150 subjects reserved for validation and testing purposes. Only the pre-contrast T1-weighted sequence was selected for subsequent analysis.
All cases underwent initial annotation using the Brain Extraction Tool (BET) from FSL~\citep{jenkinson2005bet2} and were further manually refined by a trained neuroimaging analyst from the Sydney Neuroimaging Analysis Centre. 
To eliminate any potential manual labeling errors (displayed as unexpected spikes on the mask boundaries), a morphological open operation was applied to all masks. 

Similar to \BrData, another internal dataset \MSData was curated to evaluate the performance of MS lesion segmentation. We followed the same approach of selecting unique subjects, resulting in a dataset of 152 subjects for training. To better evaluate the performance, the MSSEG-2016 dataset~\citep{commowick2021multiple} was used for validation (15 subjects) and testing (38 subjects). 
The labels in the \MSData dataset were annotated on registered T1-weighted and FLAIR images with a slice thickness of 1 mm (\ie, the axial axis slice thickness was resampled to 1mm while maintaining high resolutions in other dimensions) using a pre-configured MS lesion segmentation tool ~\citep{liu2023multiple,barnett2023ai}. These annotations underwent manual scrutiny and modification by two qualified neuroimaging analysts.
Given the enhanced visibility of MS lesions in FLAIR images~\citep{filippi2019assessment}, we chose to use FLAIR images and their corresponding annotations for further experiments.

\begin{table*}[t]
\centering
\caption{DSC scores and the estimated $\S/\V$ ratio after conducting the same Markov process to introduce acceptable noise. Only the average and standard deviation (SD) across all 10 trials were shown.}
\label{tab4:dice}
\begin{tabular}{c|ccc|ccc}
\thickline
\multirow{2}{*}{\# of Cases} 
& \multicolumn{3}{c|}{DSC Average $\pm$ SD} 
& \multicolumn{3}{c}{$\S/\V$ Average $\pm$ SD} \\ \cline{2-7} 
& \multicolumn{1}{c|}{\BrData} 
& \multicolumn{1}{c|}{BraTS} 
& \multicolumn{1}{c|}{\MSData} 
& \multicolumn{1}{c|}{\BrData} 
& \multicolumn{1}{c|}{BraTS} 
& \MSData 
\\ \hline
5      
& \multicolumn{1}{c|}{0.9769 $\pm$ 5.35e-4} & \multicolumn{1}{c|}{0.8005 $\pm$ 0.0504} & \multicolumn{1}{c|}{0.6254 $\pm$ 0.0277} 
& \multicolumn{1}{c|}{0.0476 $\pm$ 7.86e-4} & \multicolumn{1}{c|}{0.1635 $\pm$ 0.0284} & \multicolumn{1}{c}{0.7769 $\pm$ 0.0976} 
\\ \hline
10
& \multicolumn{1}{c|}{0.9769 $\pm$ 2.35e-4} & \multicolumn{1}{c|}{0.8030 $\pm$ 0.0254} & \multicolumn{1}{c|}{0.6164 $\pm$ 0.0177} 
& \multicolumn{1}{c|}{0.0478 $\pm$ 6.09e-4} & \multicolumn{1}{c|}{0.1653 $\pm$ 0.0184} & \multicolumn{1}{c}{0.6838 $\pm$ 0.0705} 
\\ \hline
20  
& \multicolumn{1}{c|}{0.9769 $\pm$ 2.11e-4} & \multicolumn{1}{c|}{0.8091 $\pm$ 0.0292} & \multicolumn{1}{c|}{0.6262 $\pm$ 0.0085} 
& \multicolumn{1}{c|}{0.0481 $\pm$ 4.94e-4} & \multicolumn{1}{c|}{0.1530 $\pm$ 0.0123} & \multicolumn{1}{c}{0.7091 $\pm$ 0.0547} 
\\ \hline
30 
& \multicolumn{1}{c|}{0.9768 $\pm$ 1.49e-4} & \multicolumn{1}{c|}{0.8064 $\pm$ 0.0205} & \multicolumn{1}{c|}{0.6194 $\pm$ 0.0106} 
& \multicolumn{1}{c|}{0.0481 $\pm$ 3.74e-4} & \multicolumn{1}{c|}{0.1555 $\pm$ 0.0095} & \multicolumn{1}{c}{0.6911 $\pm$ 0.0344} 
\\ \hline
50    
& \multicolumn{1}{c|}{0.9769 $\pm$ 1.11e-4} & \multicolumn{1}{c|}{0.8061 $\pm$ 0.0183} & \multicolumn{1}{c|}{0.6240 $\pm$ 0.0096} 
& \multicolumn{1}{c|}{0.0479 $\pm$ 2.16e-4} & \multicolumn{1}{c|}{0.1556 $\pm$ 0.0088} & \multicolumn{1}{c}{0.6945 $\pm$ 0.0211} 
\\ \hline
75 
& \multicolumn{1}{c|}{0.9769 $\pm$ 1.03e-4} & \multicolumn{1}{c|}{0.8097 $\pm$ 0.0104} & \multicolumn{1}{c|}{0.6211 $\pm$ 0.0047} 
& \multicolumn{1}{c|}{0.0481 $\pm$ 1.80e-4} & \multicolumn{1}{c|}{0.1556 $\pm$ 0.0045} & \multicolumn{1}{c}{0.6852 $\pm$ 0.0167} 
\\ \hline
100
& \multicolumn{1}{c|}{0.9769 $\pm$ 0.93e-4} & \multicolumn{1}{c|}{0.8027 $\pm$ 0.0116} & \multicolumn{1}{c|}{0.6213 $\pm$ 0.0036} 
& \multicolumn{1}{c|}{0.0479 $\pm$ 1.55e-4} & \multicolumn{1}{c|}{0.1548 $\pm$ 0.0048} & \multicolumn{1}{c}{0.6975 $\pm$ 0.0157} 
\\ \thickline
\end{tabular}
\end{table*}

For brain tumor segmentation, the publicly available BraTS dataset~\citep{menze2014multimodal} was used. It consists of 1251 cases, each with consensus labels for tumor subregions, including contrast-enhancing tumor (ET), non-enhancing tumor core (NEC), and peritumoral edema (ED). Among the various image series provided in the dataset (such as pre and post-contrast T1-weighted and T2-weighted images), we exclusively employed T1-weighted post-contrast images and focused on the combination of NEC and ED associated with this image sequence, namely represented as the tumor core (TC) in previous studies~\citep{choi2023single,pedada2023novel}, to develop the model. Similar to the split of~\citep{chen2022letcp}, we used 1000 for training, and the remaining 251 cases for validation (51) and testing (200).

\subsection{Experimental Settings}~\label{sec4:steps}
To ensure uniformity and suitability for our experiments, we pre-processed images in all three datasets, including the raw MRI scans and annotations, by using a standardized pre-processing pipeline. The common pre-processing pipeline included the following steps unless otherwise specified:

\begin{enumerate}
    \item N4 Correction~\citep{tustison2010n4itk}: This step was implemented to correct inhomogeneities in image intensity that can arise during image acquisition.
    \item Normalized Sampling: Voxel sizes across all images were normalized to $1 \times 1 \times 1 \text{mm}^3$. The data in the BraTS dataset had already undergone this processing by the dataset authors.
    \item Brain Extraction: To focus exclusively on the relevant brain regions, we performed brain extraction for all three tasks except for brain segmentation. 
    This operation was executed using our in-house brain extraction tool from \iqs~\citep{barnett2023ai} on \MSData. For the BraTS dataset, the necessary brain extraction had already been completed by the dataset authors.
\end{enumerate}

Additionally, we calculated the coordinates of bounding boxes that envelope the ROIs on the pre-processed images and masks. For each dimension of the bounding box, we determined the minimum number of patches required to cover the bounding box using predefined parameters such as the boundary size ($d_b$), overlap size ($d_o$), and patch size ($d_p$) as the described \PS method outlined in Section~\ref{sec4:reps}. In our experiments, the boundary size ($d_b$), overlap size ($d_o$) and patch size ($d_p$) were set to 16, 16 and 64, respectively.

For our experiments, we constructed subsets in an incremental manner with the number of cases increasing step by step. In this setup, data were uniformly incremented at the case level, and all patches from the newly incremented cases were integrated into the training dataset for that round. The data from previous rounds were retained for subsequent rounds to avoid uncertainty caused by different cases. 

These subsets were trained on neural networks based on the 3D U-Net architecture~\citep{ronneberger2015u,cciccek20163d}. For each task, the optimal U-Net models and the hyperparameters were selected when the networks trained on the entire dataset reached the highest performance on the test data. Then, the model structures and hyperparameters were reserved for other experiments on subsets with reduced data sizes. For each task, we conducted three rounds of training, with subsets randomly resampled from the overall dataset in each round.

We have tested our pipeline to evaluate further the effectiveness and robustness under different settings, including:
\begin{itemize}
    \item Effectiveness of \DT in estimating an achievable level of DSC by comparing with state-of-the-art results,
    \item Effectiveness of \PS in controlling the steadiness of patch selection while maintaining acceptable performance,
    \item Estimation of required data in number of cases and ROIs,
    \item Prediction of required data based on limited data.
\end{itemize}

\section{Results}
\subsection{Thresholds Decided by DSC}\label{sec4:exp-thresh}
We conducted experiments based on Section~\ref{sec4:markov} to identify the acceptable DSC for all three tasks.
For each dataset, several numbers of cases were randomly sampled from the whole dataset and repeated for 10 trials. In each trial, two steps of dilation and two steps of erosion were introduced with both $\mu_1$ and $\mu_2$ set to 0.5. The resulting averages and standard deviations related to each number of cases are presented in Table~\ref{tab4:dice}.

Table~\ref{tab4:dice} clearly illustrates that different tasks exhibit varying DSC scores, even when only boundary modifications were introduced. 
For each specific task, the expected acceptable DSC score can be determined, which should be slightly higher than the determined mean value in Table~\ref{tab4:dice}. For instance, \BrData requires an expected DSC score of approximately 0.977, while BraTS requires an expected DSC score of around 0.810, and \MSData requires an expected DSC score of roughly 0.630. 
The expected DSC scores are related to $\S/\V$ ratio in both Table~\ref{tab4:dice} and Figure~\ref{fig4:integral}. When considering the average sizes of ROIs in the three tasks (0.11ml for MS lesions, 23.05ml for enhancing brain tumors, and 1295.67ml for brains), the expected DSC scores are related to the average ROI sizes, that is, the smaller ROI size may lead to smaller expected DSC scores. 

Furthermore, as the number of cases increases in Table~\ref{tab4:dice}, the expected DSC score stabilizes, as evidenced by the decreasing standard deviation across rows. In general, it can be summarized that a range of 30 to 50 cases allows for confident calculation of expected DSC scores.

When compared with the performance reported in recently published papers, the calculated DSC scores could be further justified. In~\citep{sarica2023dense}, the recent performance on MSSEG-2016 has been reported to be 0.6727. In the recent work of~\citep{choi2023single} and~\citep{yu2023unest}, the DSC for TC has been reported to be 0.8990 using both post-contrast T1-weighted and FLAIR images, 0.9170 using all image types, and 0.8582 when only post-contrast T1-weighted images were used. The performance of brain segmentation was reported as 0.99 in~\citep{teng2023automated}. All the reported values are marginally better than the DSC scores achieved by \DT, which further proves the effectiveness of \DT.

It is also worth noting that these calculated DSC values are only determined by the ROIs of collected data and are independent of the segmentation model or training process. They serve as expected acceptable thresholds for each task, contributing to determining whether the training data are sufficient. When more data are introduced in training a more advanced model, the model's performance can obviously surpass these thresholds.

\subsection{Required Data}\label{sec4:exp-required}
\begin{figure}[h]
\begin{subfigure}{0.15\textwidth}
  \centering
  \includegraphics[width=\linewidth]{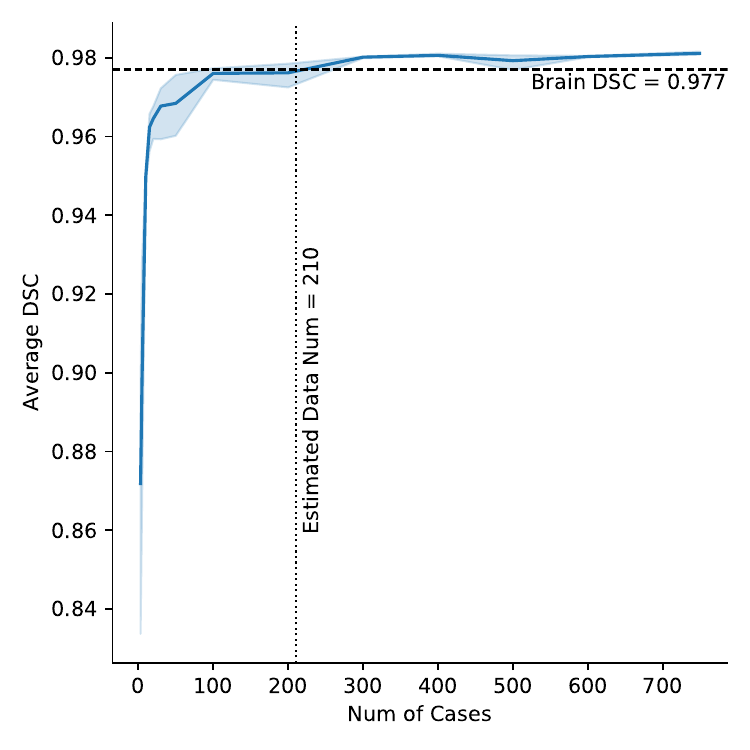}
  \caption{brain extraction}
\end{subfigure}
\begin{subfigure}{0.15\textwidth}
  \centering
  \includegraphics[width=\linewidth]{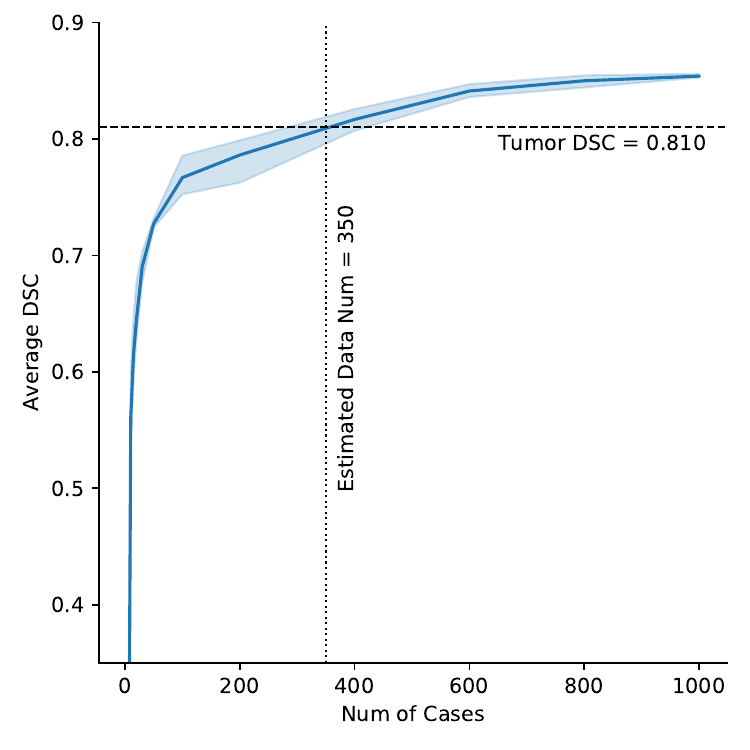}
  \caption{tumor segmentation}
\end{subfigure}
\begin{subfigure}{0.15\textwidth}
  \centering
  \includegraphics[width=\linewidth]{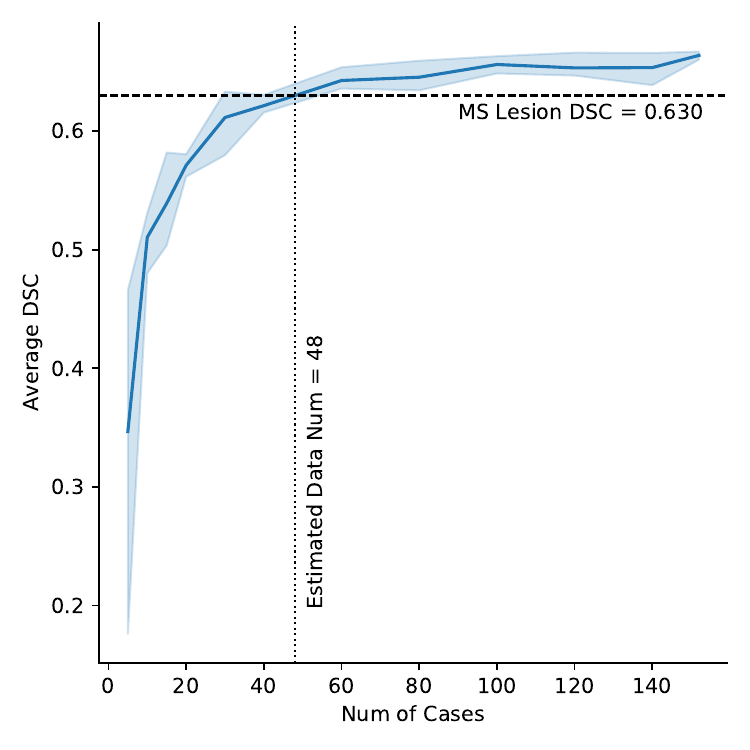}
  \caption{lesion segmentation}\label{fig4:lesion-1}
\end{subfigure}
\caption{The performance of three tasks (brain extraction, tumor segmentation, and lesion segmentation) at different numbers of training data. The black dashed line marks the expected DSC score determined by \DT, and the estimated required data numbers are shown at the crossing point.}
\label{fig4:result}
\end{figure}

Experiments on three tasks were conducted according to \PS method in Section~\ref{sec4:reps} and subsets settings in Section~\ref{sec4:steps}.
Figure~\ref{fig4:result} shows the performance changes according to different numbers of data used for training, averaged by three trials for each step in each of the tasks.

It can be observed that, as data accumulated, the performance of the related neural network models increases. The performance initially accelerates and tends to plateau once the number of cases reaches a specific number. This can be observed in all three tasks, regardless of the sizes of ROIs.

Using the DSC scores estimated using \DT in Section~\ref{sec4:exp-thresh}, the estimated required data for the three tasks can be determined, including 210 cases for brain segmentation, 350 cases for tumor segmentation, and 48 cases for lesion segmentation. The number of cases marks the minimum number of cases required to achieve a satisfactory performance, an important reference for data centers that may wish to participate in deep neural network studies, especially those that are performed in a federated learning environment. 
Specifically, clarity with respect to the related time and financial costs associated with collection and annotation of this (reference) number of cases is likely to influence their decision to participate in such studies.

\subsection{Required ROIs}
Similar to the experiments described in Section~\ref{sec4:exp-required}, we can further analyze how performance changes with regard to the ROIs of each task.

\begin{figure}[h]
\begin{subfigure}{0.15\textwidth}
  \centering
  \includegraphics[width=1.1\linewidth]{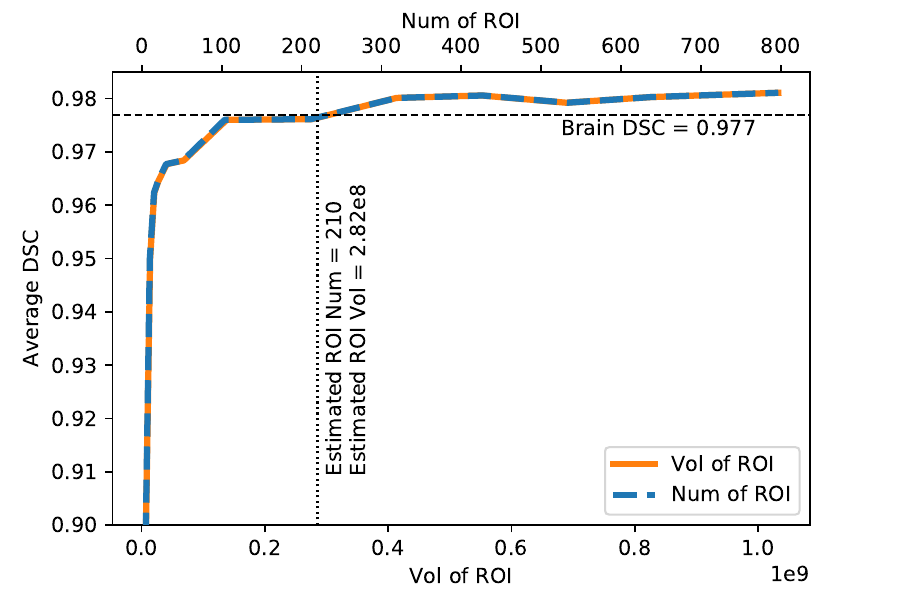}
  \caption{brain extraction}
\end{subfigure}
\begin{subfigure}{0.15\textwidth}
  \centering
  \includegraphics[width=1.1\linewidth]{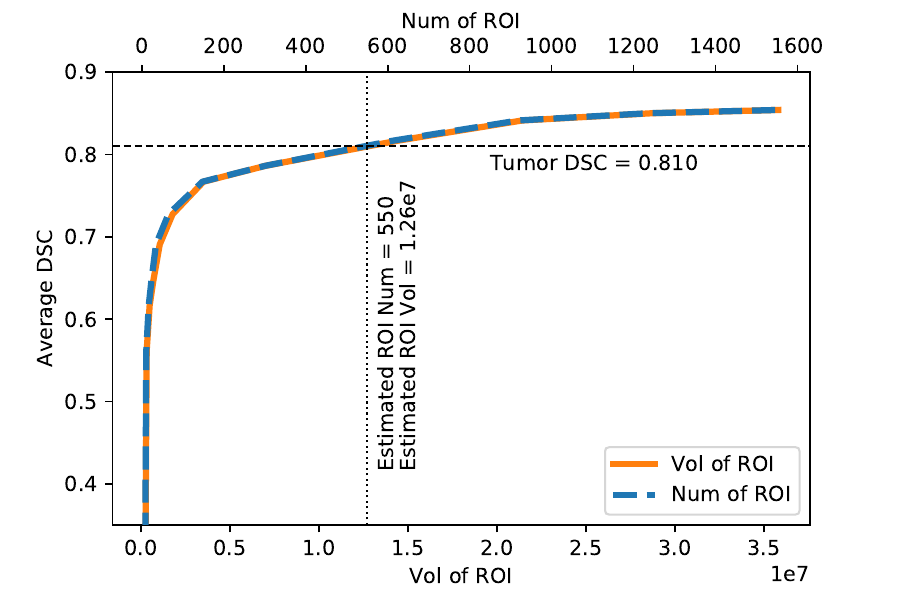}
  \caption{tumor segmentation}
\end{subfigure}
\begin{subfigure}{0.15\textwidth}
  \centering
  \includegraphics[width=1.1\linewidth]{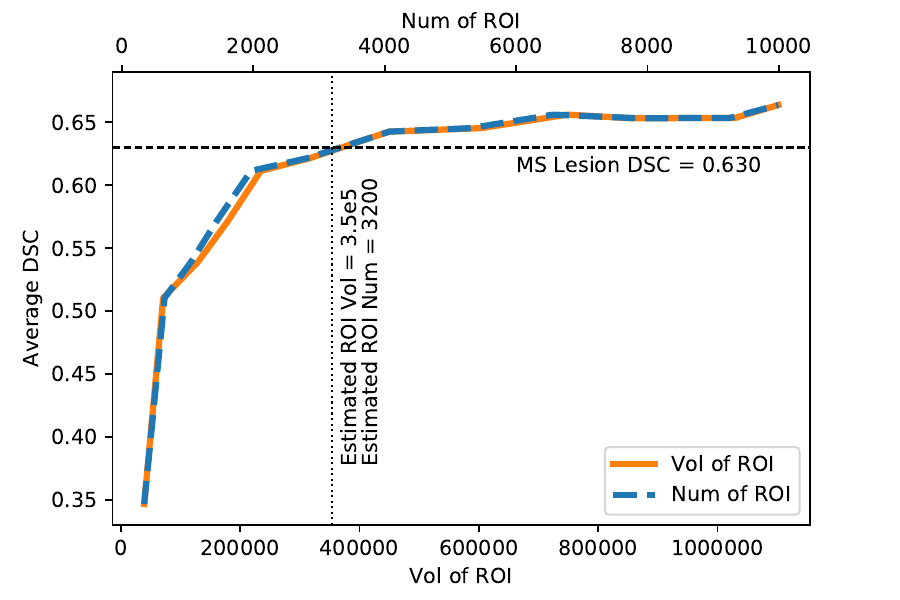}
  \caption{lesion segmentation}\label{fig4:lesion-2}
\end{subfigure}
\caption{The performance of three tasks (brain extraction, tumor segmentation, and lesion segmentation) at different numbers of training ROIs. The black dashed line marks the expected DSC score determined by \DT, and the estimated required ROI numbers are shown at the crossing point.}
\label{fig4:roi}
\end{figure}

Figure~\ref{fig4:roi} demonstrates the performance changes of models as the volumes and numbers of ROIs used in the training increase. 
It can be observed that from large ROI to small ROI, the required number of ROIs increases (210, 550, and 3200 ROIs for three tasks), and the total volumes of ROIs used for training decreases (2.82e8ml, 1.26e7ml, and 3.50e5ml for three tasks).
 
This observation suggests that when the target ROIs are small, it is beneficial to use more ROIs to train a better model. Small ROIs, such as MS lesions, may include more severe variance compared with large ROIs (such as the whole brain mask). 
Compared with the findings revealed in Figure~\ref{fig4:result}, the DSC is more closely associated with the characteristics of ROIs rather than the number of cases. Therefore, it is advisable to prioritize case selection with a focus on those with more pathology or a higher disease burden for training.

It is also hypothetical that the increase in the total volume of ROIs might lead to better DSC scores, as the achievable DSC thresholds decided by \DT for brain, tumor, and lesion segmentation tasks decrease and the involved ROI volumes also decrease. 
However, for tasks involving small ROIs such as MS lesion segmentation, it may not be feasible to collect a dataset that is comparable in size to the other two datasets shown in Figure~\ref{fig4:roi}, which could require another 100$\times$ to 1000$\times$ more ROIs, resulting in over 40,000 MS cases. 
This finding provides additional insights for project owners, emphasizing the importance of carefully balancing performance and cost when making decisions about data collection and model training.

\subsection{Prediction of the Required Data}\label{sec4:exp-predict}

\begin{figure}[h]
\begin{subfigure}{0.15\textwidth}
  \centering
  \includegraphics[width=1.1\linewidth]{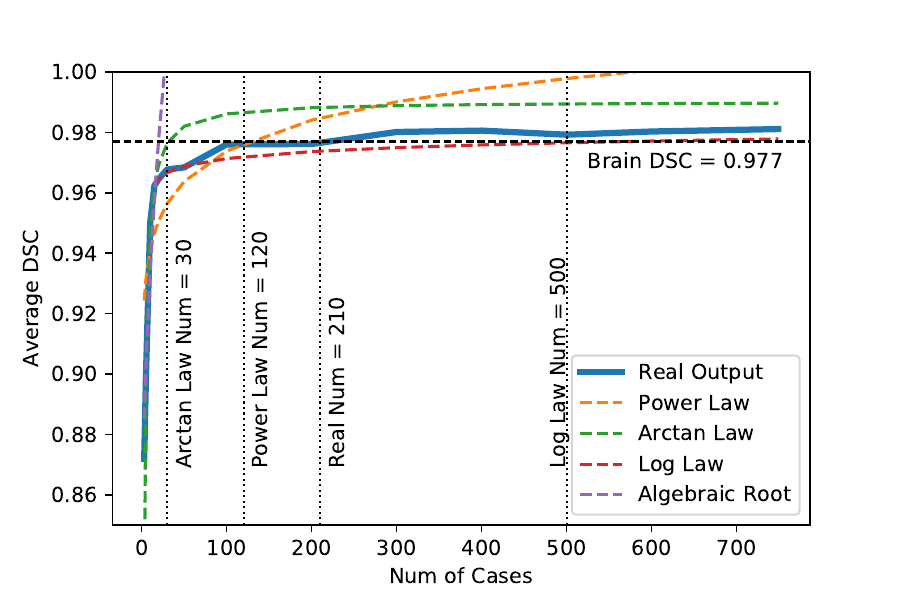}
  \caption{brain extraction}
\end{subfigure}
\begin{subfigure}{0.15\textwidth}
  \centering
  \includegraphics[width=1.1\linewidth]{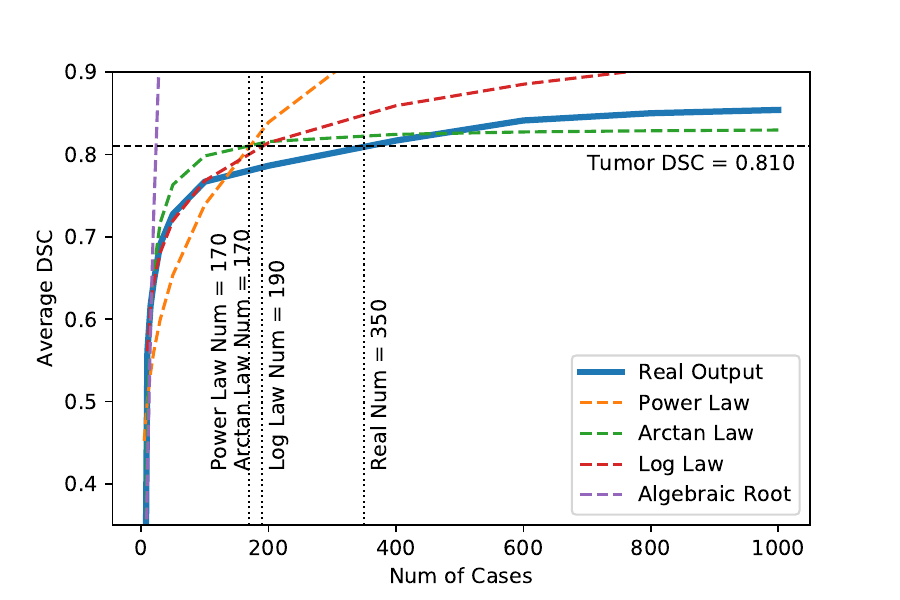}
  \caption{tumor segmentation}
\end{subfigure}
\begin{subfigure}{0.15\textwidth}
  \centering
  \includegraphics[width=1.1\linewidth]{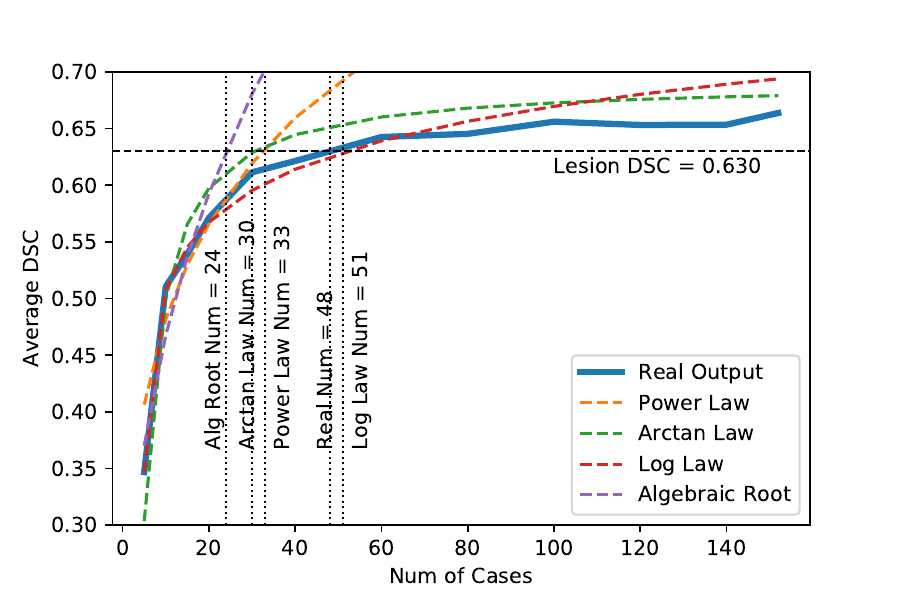}
  \caption{lesion segmentation}\label{fig4:lesion-3}
\end{subfigure}
\caption{The prediction of required data by different regression functions, shown in dashed lines in each image.}
\label{fig4:predict}
\end{figure}

It is important to predict the required data, even when only a relatively small number of cases have been collected. Based on previous work~\citep{mahmood2022much}, several functions can be used to represent the relationship between expected performance and the required data and make predictions, including:

\begin{equation}
\begin{aligned}
    \text{Power Law:} \qquad & \theta_1 n^{\theta_2} + \theta_3, \\
    \text{Arctan Law:} \qquad & \frac{200}{\pi} \arctan (\theta_1\frac{\pi}{2} n + \theta_2) + \theta_3, \\
    \text{Logarithmic:} \qquad & \theta_1 \log (n + \theta_2) + \theta_3, \\
    \text{Algebraic Root:} \qquad & \frac{100n}{(1+ | \theta_1 n |^{\theta_2})^{1/\theta_2}} + \theta_3,
\end{aligned}\label{eq4:laws}
\end{equation}
where the calculation results are the target DSC in our setting with regards to $n$ as the number of cases, and $\bm{\theta} = \{\theta_1, \theta_2, \theta_3\}$ are learnable parameters. Based on the algorithm provided in~\citep{mahmood2022much}, we calculate the prediction curves using the first 20 cases of each task (randomly selected 5, 10, 15, and 20), and the predictions are shown in Figure~\ref{fig4:predict}.

It can be observed that prediction of the required number of cases can be achieved after the inclusion of only the first 20 cases for all tasks, without regard to the sizes of ROIs, and Logarithm Law is relatively more suitable and realistic to predict the number of requirements for patch-based segmentation tasks although none of the functions could provide very accurate prediction. As previously pointed out by~\citep{mahmood2022much}, the prediction gaps could be mitigated with more collected data and accumulated experiments.

\subsection{Effectiveness of \PS}
\begin{figure}[h]
\begin{subfigure}{0.15\textwidth}
  \centering
  \includegraphics[width=0.8\linewidth]{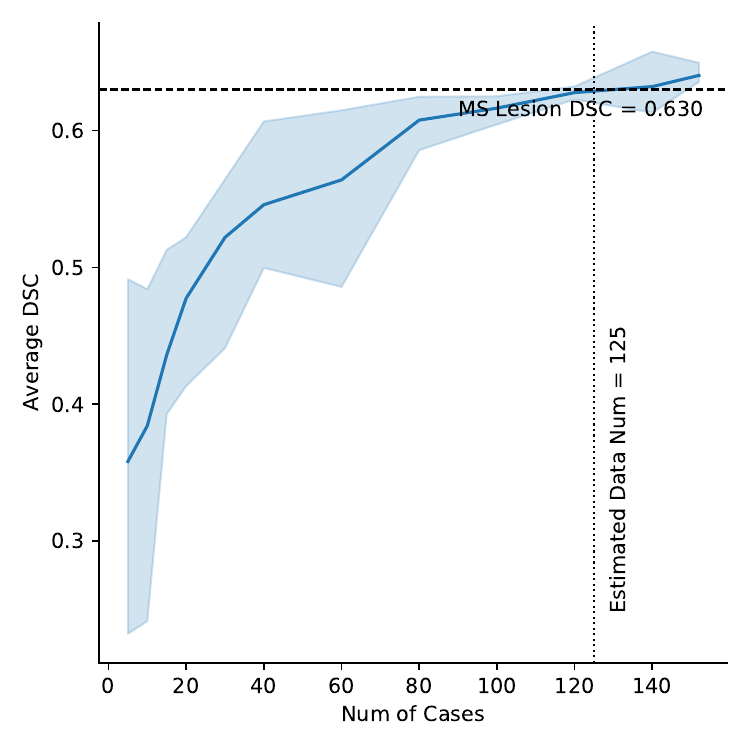}
  \caption{performance vs cases}
\end{subfigure}
\begin{subfigure}{0.15\textwidth}
  \centering
  \includegraphics[width=1.1\linewidth]{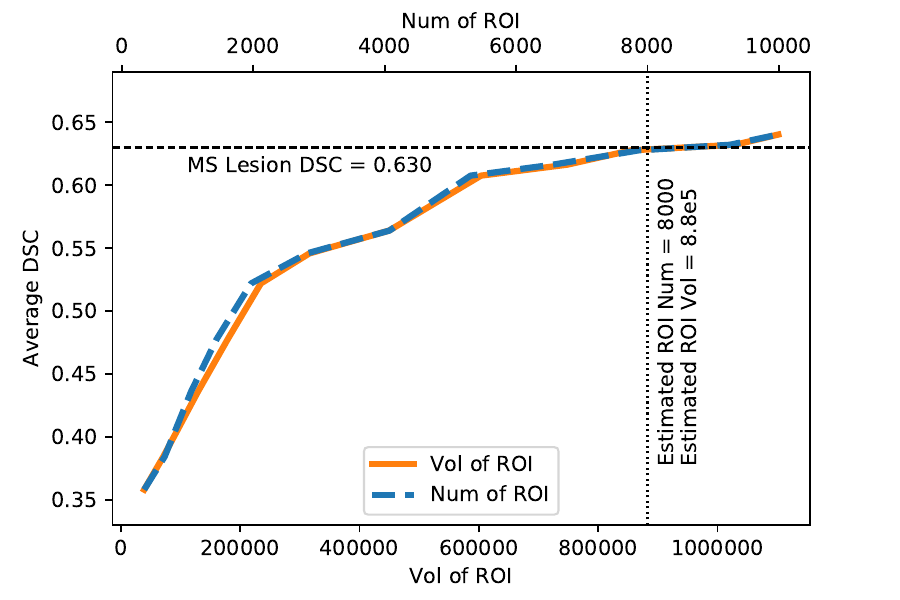}
  \caption{performance vs ROIs}
\end{subfigure}
\begin{subfigure}{0.15\textwidth}
  \centering
  \includegraphics[width=1.1\linewidth]{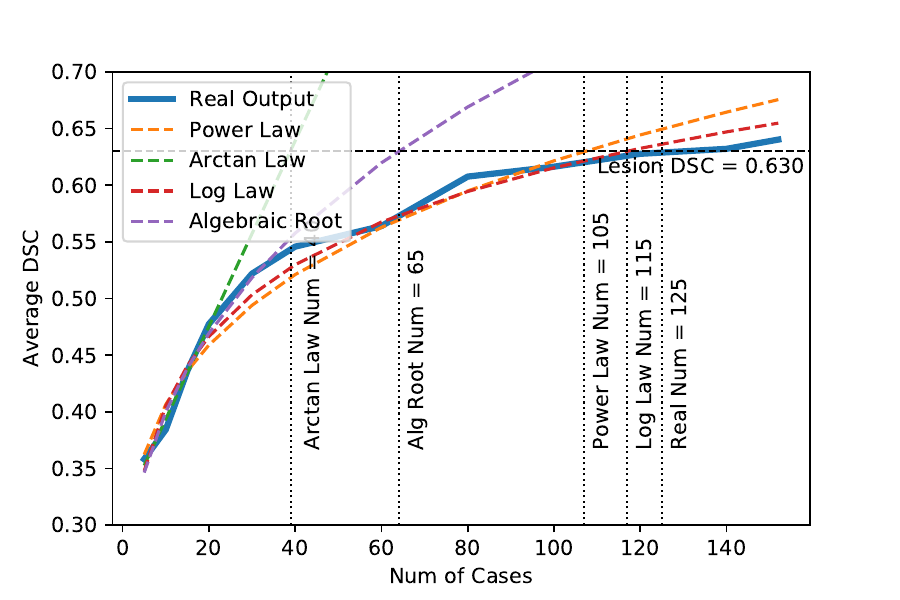}
  \caption{predicted cases}
\end{subfigure}
\caption{The performances of the baseline patch selection method on the MS lesion segmentation task.}
\label{fig4:baseline}
\end{figure}

The motivation of \PS is to maintain the performance of the network while regulating the contribution of each case to the model training process, which is intended to predict the data requirements more accurately. 
This could be further validated by comparing performance with a baseline method that randomly selects an equal number of patches from each case during training. The same experiments were conducted on \MSData and MSSEG-2016, and the performance of the baseline method was shown in Figure~\ref{fig4:baseline}, which can be compared with the performance of \PS in Figures~\ref{fig4:lesion-1},~\ref{fig4:lesion-2} and~\ref{fig4:lesion-3}, respectively.

Figure~\ref{fig4:baseline} illustrates that the baseline method shows a steadily increasing performance as the number of training cases grows, although the fluctuation among different trials is more severe compared with the performance shown in Figure~\ref{fig4:lesion-1}. 
This phenomenon may be caused by model underfitting associated with the random patch selection process. In this process, patches were selected randomly and may not accurately represent the characteristic patterns of the cases. 
The experiments also emphasize the influence of the patch selection method on the estimation of cases required, in which the baseline method may incur a vastly higher burden of cases. According to Figure~\ref{fig4:baseline}, it requires a considerably larger number of cases to reach the DSC threshold set by \DT, which is 125 cases of in total 8.80e5ml ROIs, almost 2.5 times of the results of the method based on \PS.

Compared with the baseline method, \PS shows a favorable balance by preserving model performance while providing more accurate estimates of the required number of cases.

\section{Discussion}

We introduce a novel pipeline aimed at estimating and evaluating requisite sample sizes for training satisfactory patch-based segmentation models. This involves the implementation of the \DT method, which estimates acceptable DSC scores, and the \PS method, designed to conduct experiments that standardize the contribution of cases to model training.

By comparing the proposed DSC scores of the three tasks with values calculated through Equation~\ref{eq4:integral}, a strong correlation between the DSC values and the $\S/\V$ ratio is evident. The state-of-the-art results slightly surpass the proposed DSC, affirming the effectiveness of \DT. Additionally, the \PS method's efficacy is confirmed by comparing with the random selection of ROIs, and the proposed method ensured a balanced contribution of different cases while preserving overall and steadily increasing performance as training samples accumulated.

Notably, Equation~\ref{eq4:integral} and our experiments unveil a discernible correlation between expected DSC performance and the $\S/\V$ ratio, displaying an approximately negative linear relationship. A comparative analysis across brain, tumor, and lesion segmentation tasks suggests that superior performance is associated with more ROIs across cases, while achieving higher DSC scores proves challenging for tasks with smaller ROIs.

\subsection{Selection of DSC Calculation Methods}
The Dice similarity coefficient (DSC), originally introduced by Dice in~\citep{dice1945measures}, remains a dominant metric for evaluating segmentation tasks. 
However, several inherent issues have been identified concerning DSC scores. In the visualization examples presented by~\citep{reinke2021common} 
, the authors analyzed several scenarios in which the DSC may inadequately reflect the real performance of the models being studied. This is especially noticeable in the context of significant heterogeneity in the size, volume, or shape of ROIs, false negative outliers, and false positive structures.

Advanced DSC-based scores have been proposed to address some of these limitations. 
In~\citep{shamir2019continuous}, a weighted DSC was introduced in which a continuous similarity coefficient was calculated based on probability masks instead of binary masks. The robustness of this new metric was enhanced through the introduction of a weighting factor to balance the contributions of the ground-truth mask and the predicted probability mask.
In~\citep{wang2020improved}, a novel DSC-based metric was suggested to balance large and small lesions when calculating the DSC. Furthermore, \cite{carass2017longitudinal} recommended using the DSC in conjunction with other metrics to provide a more comprehensive assessment of model performance.

\begin{figure}[t]
\centerline{\includegraphics[width=0.9\columnwidth]{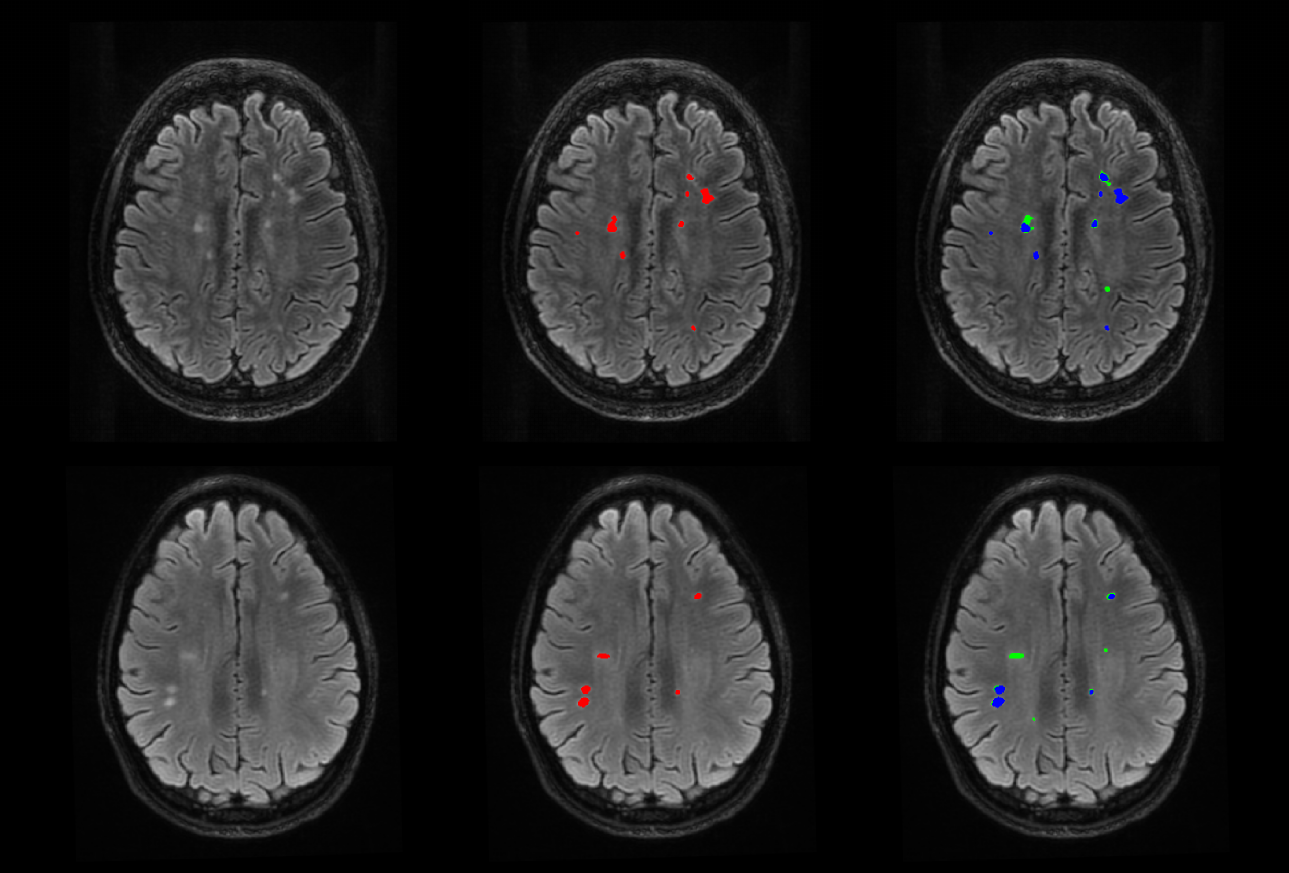}}
\caption{An example of label uncertainty from different data centers. Each row represents one single case, and the three columns represent the original slices, consensus labels, and noisy labels respectively. The annotations in red are the consensus of several expert image analysts and the annotations in blue and green are from different annotators and are intended to illustrate the underestimation or overestimation of the lesions correspondingly. Better viewed in color.}
\label{fig4:example}
\end{figure}

In the present work, we have opted to use the DSC as a performance metric given its widespread adoption in the majority of segmentation network development. As the ground truth for evaluation is mostly produced by human experts, it is safe to assume that most ground-truth labels inherently contain variations at the mask border, where voxel intensities vary the most and thereby impact the decision-making during annotation (as shown in Figure~\ref{fig4:example} for lesions).  
We utilized a Markov process in \DT to simulate randomness at the annotation boundaries and model the uncertainty according to specific tasks with different sizes and shapes of ROIs. We also used this estimated DSC score as the threshold for accepting the performance and the number of cases being used in our experiments.

\subsection{Understanding the DSC Thresholds}
The primary purpose of designing \DT is to simulate the uncertainty present in ground-truth masks, particularly along borders, which can often be introduced by human annotators.

As illustrated in Figure~\ref{fig4:example}, the uncertainty arises due to the partial volume effect, a phenomenon where a single voxel contains a mixture of different tissue types due to the limited spatial resolution of the imaging technique~\citep{ballester2002estimation}. This leads to variations in voxel contrast at the borders of target tissues or pathologies. For human annotators, these regions with manual analysis errors and inherent variances result in segmentation uncertainty.

As modeled by \DT, these uncertain regions typically cover just one to two voxels along the border. Therefore, these marginal voxel changes do not significantly contribute to lowering the DSC for larger ROIs or ROIs close to a sphere. Conversely, when the ROI is comparable in size to the marginal uncertainty, even slight variances can lead to a significant fall in DSC, as demonstrated in Table~\ref{tab4:dice} for MS lesions.

Therefore, inferior performance of the DSC when compared with the ground truth produced by analysts may not necessarily indicate that the model requires further improvement. Rather, it is likely that the model has already reached acceptable performance for robust segmentation of the tissues based on the given data, and deviation from the ideal DSC (which is 1) can be attributed to manual variances introduced during the ground-truth label creation. Based on this assumption, the threshold established by \DT could be used to set the expectations of DSC for specific tasks and avoid unnecessary data preparation and effort to optimize a model. 

\subsection{Selection of \DT Hyperparameters}\label{sec4:hyper}

\begin{figure}[t]
\centerline{\includegraphics[width=0.9\columnwidth]{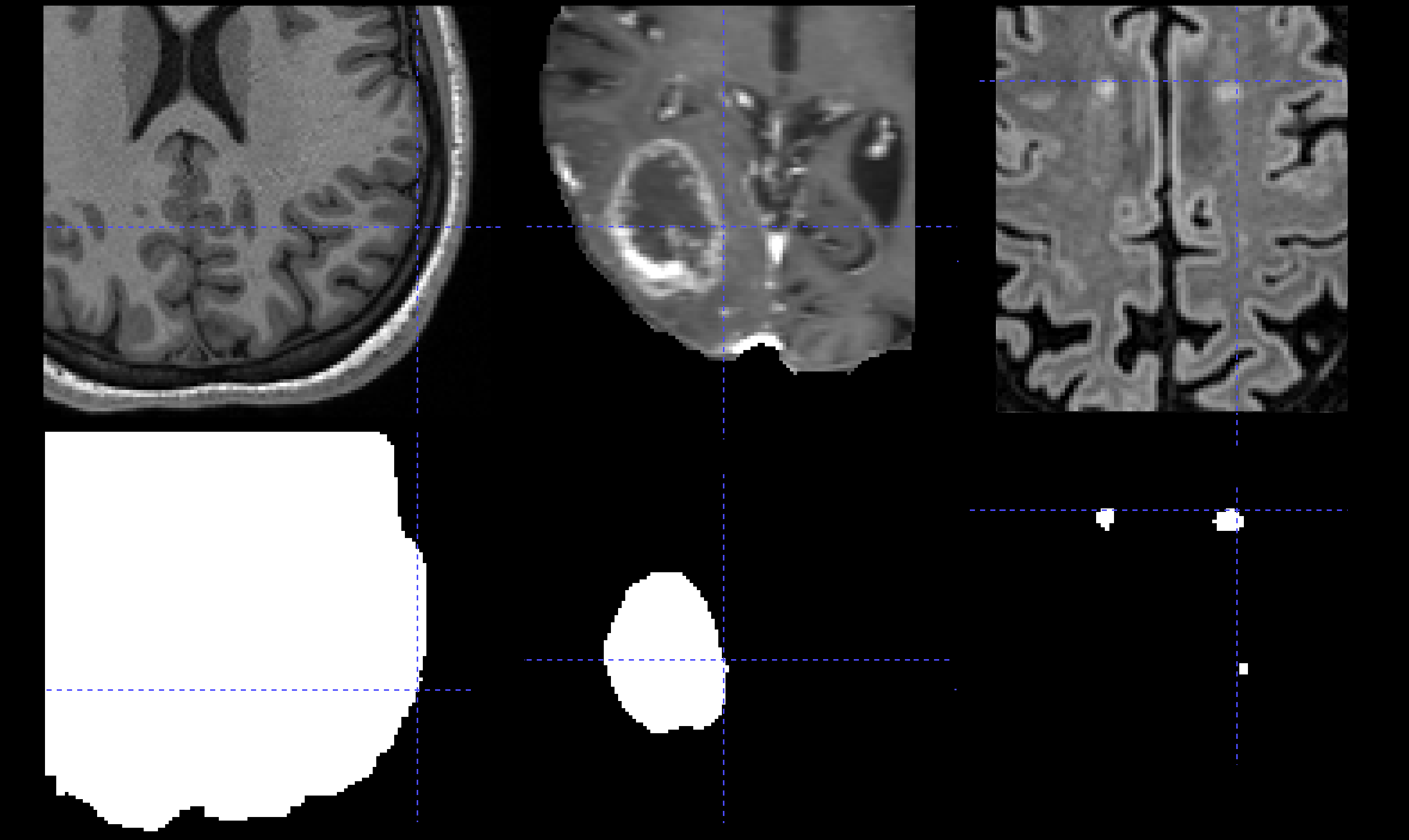}}
\caption{An example of ROI boundaries for the three types of data. From left to right: brain mask in T1-weighted image, tumor core mask in post-contrast T1-weighted image, and MS lesion mask in FLAIR image. The boundaries of brain masks and lesion masks are relatively more distinctive than the boundaries of tumor core masks. Better viewed by zooming in.}
\label{fig4:boundary}
\end{figure}

The DSC estimation output may be influenced by the hyperparameters of \DT, which encompass the probability of boundary change (\ie, $\mu_1$ and $\mu_2$, both set to 0.5 in our experiments), the number of steps involved in the Markov process for boundary changes (specifically, two dilation and two erosion steps, as illustrated in Figure~\ref{fig4:markov}), and the size of boundary changes (\ie, the size of morphological kernels in each step of the Markov process). It is evident that higher probabilities, more steps, and larger morphological kernel sizes may potentially result in lower DSC thresholds.

It is crucial to emphasize that our method, \DT, is solely related to the shape and size of the target ROIs and the extent of blur on ROI boundaries in the original images, and it is independent of the model structure used for training and prediction. 
Based on our analysis of ROI boundaries in the original images (examples shown in Figure~\ref{fig4:boundary}), we observed that the boundaries of the brain and lesions are more distinguishable than those of the tumor core. Consequently, we have opted to set the random change of boundaries to a maximum of 3 voxels for BraTS, and the maximum boundary change to 2 voxels for the other two datasets, leading to the results listed in Table~\ref{tab4:dice}. The differences in the maximum boundary change may also influence the estimation of \DT as shown by two different estimations for tumor segmentation in Figure~\ref{fig4:integral}, and the lower estimation for the tumor segmentation task is proved to be more realistic judging from the experiments and state-of-the-art performances.

\subsection{Influence of the Learning Curves}
Our experiments show the learning curves and the projected essential data requirements proposed by \DT and \PS methods. The learning curves were based on a 3D U-Net based model, and it is important to note that the choice of model design and the utilization of alternative sophisticated approaches can significantly impact both the learning curves and the required number of cases.

For example, the influence of noise labeling on model performance, even with constant dataset size, is related to the level of label noise and is hard to predict. The method discussed in~\citep{bai2023improving} offers a potential way to tackle label noise and expedite performance gains with increasing data size.  
Additionally, domain adaptation strategies~\citep{liu2023multiple}, which leverage knowledge transfer from other datasets to address tasks with limited sample sizes, could prove different data requirements in this context. 
Likewise, the adoption of semi-supervised or unsupervised approaches may further alleviate some labeling processes, which may also potentially reduce the cost of labeling and the overall data requirements while introducing performance gains. 
When employing pre-trained models, particularly larger models, for training segmentation networks as downstream tasks, the anticipated target and data requirements can undergo significant alterations.

Furthermore, it is worth noting that all these three tasks were trained on the most distinctive sequence type. Although introducing more modalities under the same number of subjects may yield gains in the learning curve, the inclusion of additional modalities on the same subjects does not necessarily provide a comparable benefit, according to the performance reported by~\citep{choi2023single,yu2023unest}. This observation may be attributed to the limited information gain from extra modalities. 

Considering the diverse conditions and variants, predicting data requirements with a limited dataset~\citep{mahmood2022much} and a constrained number of experiments (refer to Figure~\ref{fig4:predict}) proves challenging. 
As a result, it is recommended to follow the methodology of \PS to progressively identify the data requirement as both training data and experiments accumulate.

\section{Conclusion}
``How much data are required?'' is perhaps the first question asked to a machine learning engineer when commencing a development task. Many machine learning experts will answer ``the more the better'', and others may provide an intuitive number based on previous experience. While neither response is incorrect, the former may have an unrealistic financial burden, sometimes in the realm of millions of dollars, which may not be necessary for the performance required; and the latter is subjective and may pose a high risk of either over or under-budgeting data preparation costs.

The work described in this paper is, to our knowledge, the first to propose an end-to-end strategy to predict both performance expectations and the requisite size of the associated training datasets based on the given task, even before model development begins. 
The strategy we propose is important to real-world applications. With a newly defined task, sample datasets and annotations can provide information on the expected acceptable performance, and with incremental training datasets, the requirement of data size can be inferred while the model is being developed.
Although the experiments described herein are focused on neuroimaging applications, this framework can be easily extended to other domains and, ultimately, provide evidence-based guidance for the expectation of data size required in any segmentation task.

\section*{Acknowledgment}
This work was supported in part by the Australian Department of Health under the Australian Medical Research Future Fund MRFAI000085. Besides, this project and related co-authors were supported by the Australian Research Council Grant DP200103223.

\section*{Appendix}\label{sec4:app}
The expectation of DSC can be estimated based on Equation~\ref{eq4:factor} by calculating the expectation with respect to $\mu_1$ and $\mu_2$.
Recall the expression of DSC as $$\text{DSC} = \frac{1- \mu_2 \cdot C}{1 - (\mu_2/2 - \mu_1/2) \cdot C}$$ where $C:=\S/\V \in (0,1)$ when considering both $\S$ and $\V$ are represented by voxels and $\mu_1,\mu_2$ are the independent uniform-distributed random variables on $[0,1]$. Take a substitution of the variables by letting 
$$
\gamma=\frac{\alpha}{\beta}, \  \alpha=1+\mu_1 C, \ \beta = 1-\mu_2 C
$$
where $\alpha \sim \mathbf{U}[1,1+C]$ and $\beta \sim \mathbf{U}[1-C,1]$ independently, and 
$$
\text{DSC}^{-1} = \frac{1}{2} (1+\gamma). 
$$
Then we obtain a joint probability density function (pdf) for $(\gamma, \beta)$, i.e.
$$
p_{\gamma,\beta}(r,b)=|det(J)|(r,b)p_{\alpha}(rb)p_{\beta}(b)
$$
where $p_{\alpha}$ is the pdf of the random variable $\alpha$ in the form of $p_{\alpha}(a)=\frac{1}{C}\mathbf{1}\{1 \leq a \leq 1+C \}$, $p_{\beta}$ the pdf of the random variable $\beta$ with $p_{\beta}(b)=\frac{1}{C}\mathbf{1}\{1-C \leq b \leq 1 \}$ and $J$ is the Jacobian matrix given by 
$$
\begin{pmatrix}
    \frac{\partial \alpha}{\partial \gamma} & \frac{\partial \alpha}{\partial \beta} \\
    \frac{\partial \beta}{\partial \gamma} & \frac{\partial \beta}{\partial \beta} \\
\end{pmatrix} 
$$
To get the marginal distribution of $\gamma$, it follows that 
\begin{equation}
\begin{aligned}
    p_{\gamma}(r) &= \frac{1}{C^2}\int _{1-C}^{1} |b| \mathbf{1}\{1 \leq rb \leq 1+C\} db \\
    &= \frac{1}{2C^2} \Bigg\{\min\left[1, \frac{(1+C)^2}{r^2}\right] 
- \max \left[(1-C)^2, \frac{1}{r^2}\right] \Bigg\}
\end{aligned}
\end{equation}

It follows that the expectation of \text{DSC} can be obtained as 

\begin{equation}
\begin{aligned}
\mathbb{E}(\text{DSC}) = \frac{1}{C^2} \int_{1-C}^{1} t^{-1} &\Bigg\{ \min\left[1, \frac{(1+C)^2}{(2t^{-1}-1)^2}\right] \\
&- \max \left[(1-C)^2, \frac{1}{(2t^{-1}-1)^2}\right] \Bigg\} dt
\end{aligned}
\end{equation}
which can be approximated by 
\begin{equation}
  \begin{array}{l}
   \mathbb{E}(\text{DSC}) \approx -0.02788364C^3 +  0.00628077C^2 \\
   \ \ \ \ \ \ \ \ \ \ \ \ \ \ \ \ \ \ -0.5016117C + 1.00008759,
  \end{array}
\end{equation}
where $C$ is the ratio $\S/\V$. 

\bibliographystyle{model2-names.bst}\biboptions{authoryear}
\bibliography{paper}

\end{document}

%% file: definitions.tex
\usepackage{etoolbox}
\usepackage{bm}

\usepackage{amsthm}
\usepackage{algorithm}
\usepackage{algorithmic}
\usepackage{subcaption}
\usepackage{xspace}
\usepackage{url}

\newcommand{\MyMapTemplatePrefixc}[4]{\expandafter#1\csname#3#4\endcsname{#2{#4}}} 
\forcsvlist{\MyMapTemplatePrefixc {\def} {\mathcal}{c}} {A,B,C,D,E,F,G,H,I,J,K,L,M,N,O,P,Q,R,S,T,U,V,W,X,Y,Z}  

\newcommand{\MyMapTemplatePrefixtb}[5]{\expandafter#1\csname#4#5\endcsname{#2{#3{#5}}}} 
\forcsvlist{\MyMapTemplatePrefixtb {\def} {\tilde}{\mathbf}{t}} {A,B,C,D,E,F,G,H,I,J,K,L,M,N,O,P,Q,R,S,T,U,V,W,X,Y,Z}  
\forcsvlist{\MyMapTemplatePrefixtb {\def} {\tilde}{\mathbf}{t}} {0,1,a,b,c,d,e,f,g,h,i,j,k,l,m,n,o,p,q,r,s,u,v,w,x,y,z}  

\newcommand{\MyMapTemplateNoPrefix}[3]{\expandafter#1\csname#3\endcsname{#2{#3}}}
\forcsvlist{\MyMapTemplateNoPrefix {\def} {\mathbf} } {0,1,a,b,d,e, f, g, h, j, k, m, n, p, q, r, u, w, x, y, z} 
\forcsvlist{\MyMapTemplateNoPrefix {\def} {\mathbf} } {A,B,C,D,E,F,G,H,I,J,K,L,M,N,O,P,Q,R,S,T,U,V,W,X,Y,Z}  

\def\etal{\textit{et~al.}\@\xspace}
\def\ie{\textit{i.e.}\@\xspace}
\def\eg{\textit{e.g.}\@\xspace}

\usepackage{booktabs} 